\documentclass{article}

\usepackage[preprint]{neurips_2021}

\usepackage[utf8]{inputenc} 
\usepackage[T1]{fontenc}    
\usepackage{hyperref}       
\usepackage{url}            
\usepackage{booktabs}       
\usepackage{makecell}       
\usepackage{amsfonts}       
\usepackage{amsmath}        
\usepackage{mathtools}
\usepackage{nicefrac}       
\usepackage{microtype}      
\usepackage{xcolor}         
\usepackage{soul}           
\usepackage{graphicx}       
\usepackage{float}          
\usepackage[ruled]{algorithm2e}
\usepackage[all]{hypcap}    

\title{Online reinforcement learning with sparse rewards through an active inference capsule}

\author{%
    Alejandro Daniel Noel\\
    Department of Cognitive Robotics\\
    Delft University of Technology\\
    \texttt{adanielnoel@gmail.com} \\
    \And
    Charel van Hoof \\
    Department of Cognitive Robotics\\
    Delft University of Technology\\
    \texttt{charel.van.hoof@gmail.com} \\
    \AND
    Beren Millidge \\
    MRC Brain Network Dynamics Unit \\
    University of Oxford \\
    \texttt{beren@millidge.name}
}

\begin{document}
\maketitle

\begin{abstract}
    Intelligent agents must pursue their goals in complex environments with partial information and often limited computational capacity. Reinforcement learning methods have achieved great success by creating agents that optimize engineered reward functions, but which often struggle to learn in sparse-reward environments, generally require many environmental interactions to perform well, and are typically computationally very expensive. Active inference is a model-based approach that directs agents to explore uncertain states while adhering to a prior model of their goal behaviour. This paper introduces an active inference agent which minimizes the novel \textit{free energy of the expected future}. Our model is capable of solving sparse-reward problems with a very high sample efficiency due to its objective function, which encourages directed exploration of uncertain states. Moreover, our model is computationally very light and can operate in a fully online manner while achieving comparable performance to offline RL methods. We showcase the capabilities of our model by solving the mountain car problem, where we demonstrate its superior exploration properties and its robustness to observation noise, which in fact improves performance. We also introduce a novel method for approximating the prior model from the reward function, which simplifies the expression of complex objectives and improves performance over previous active inference approaches.
\end{abstract}

\section{Introduction}
The field of Reinforcement Learning (RL) has achieved great success in designing artificial agents that can learn to navigate and solve unknown environments, and has had significant applications in robotics \citep{Kober2013, Polydoros2017}, game playing \citep{Mnih2015, Silver2017, Shao2019}, and many other dynamically varying environments with nontrivial solutions \citep{Padakandla2020}. However, environments with sparse reward signals are still an open challenge in RL because optimizing policies over Heaviside or deceptive reward functions such as that in the mountain car problem requires substantial exploration to experience enough reward to learn.

Recently, Bayesian RL approaches \citep{Ghavamzadeh2015} and the inclusion of novelty in objective functions \citep{Stadie2015, Burda2018, Shyam2019} have begun to explicitly address the inherent exploration-exploitation trade-off in such sparse reward problems. In parallel to these developments, active inference (AIF) has emerged from the cognitive sciences as a principled framework for intelligent and self-organising behaviour which naturally often converges with state of the art paradigms in RL (e.g., \cite{Friston2009, Friston2015d, Kaplan2018a, Tschantz2020a}). AIF agents minimize the divergence between an unbiased generative model of the world and a biased generative model of their preferences (shortly, the \textit{prior}). This objective assigns an epistemic value to uncertain states, which enables directed exploration. Because of its principled foundations and because reward functions can be seen as an indirect way of defining prior models (cf. reward shaping \citep{Ng1999}), active inference is often presented as a generalization of RL, with KL-control and control-as-inference as close ontological relatives \citep{Millidge2020a}.

\subsection{Related work}
Until recently, active inference implementations have been constrained to toy problems in theoretical expositions \citep{Friston2015d, Friston2017a, Friston2017c}. Based on the work by \cite{Kingma2014} on amortized variational inference, \cite{Ueltzhoffer2018} proposed the first scalable implementation of AIF using deep neural networks to encode the unbiased generative model and evolution strategies to estimate policy gradients from multiple parallel simulations on a GPU. Later publications proposed more efficient policy optimization schemes, such as amortized policy inference \citep{Millidge2019} and applying the cross-entropy method \citep{Tschantz2019, Tschantz2020a}. This latter work also uses an improved extension of the model free energy to future states, namely, the \textit{free energy of the expected future} \citep{Millidge2020c} (cf. divergence minimization \citep{Hafner2020}). In these papers, active inference is shown to deliver better performance than current state of the art RL algorithms on sparse-reward environments, although they use the goal states as hard-coded priors. We improve upon the model of \cite{Tschantz2020a} by demonstrating fully online learning on a single CPU core, by modeling the transition model with gated recurrent units that can capture environment dynamics over longer periods, and by learning the prior model from the (sparse) reward function through a novel reward shaping algorithm.


\section{Active inference}
The objective of an active inference (AIF\footnote{It is common to abbreviate \textit{active inference} as AIF to avoid confusion with \textit{artificial intelligence}.}) agent is to minimize surprise, defined as the negative log-likelihood of an observation, $-\ln p(y)$. However, it is often intractable to compute this quantity directly. Instead, we apply variational inference and minimize a tractable upper bound for surprisal, namely, the variational free energy (VFE) of a latent model of the world. The agent possesses an approximate posterior distribution $q(x\mid y)$, where $x$ is the latent state that is optimized to minimize the variational free energy. The parameters of this approximate posterior can be thought of as the agent's `beliefs' about its environment. The variational free energy can be written as:
\begin{equation}
  \label{eq:VFE}
  \mathrm{VFE} = \mathbb{E}_{q(x_t\mid y_t)}\left[-\ln p\left(y_t\mid x_t\right)\right] + \mathrm{D_{KL}}\left[q(x_t\mid y_t)\|\, p(x_t)\right]
\end{equation}
which is equivalent to the negative of the expectation lower bound (ELBO) used in variational inference (e.g., \citep{Attias1999, Kingma2014}).

Active inference agents select actions that are expected to minimize the path integral of the $\mathrm{VFE}$ for future states \citep{Friston2012d}. There are two common extensions of the $\mathrm{VFE}$ to account for future states, the \textit{expected free energy} \citep{Friston2015d} and the \textit{free energy of the expected future} ($\mathrm{FEEF}$) \citep{Tschantz2020a}, which we use in this work. \cite{Millidge2020c} argues that the $\mathrm{FEEF}$ is the only one consistent with \autoref{eq:VFE} when evaluated at the current time and additionally considers the expected entropy in the likelihood $p\left(y_t\mid x_t\right)$ when selecting a policy.

\subsection{Free energy of the expected future}\label{sec:feef}
The $\mathrm{FEEF}$ is a scalar quantity that measures the KL-divergence between unbiased beliefs about future states and observations and an agent's preferences over those states. The preferences are expressed as a biased generative model of the agent $\tilde{p}\left(y, x\right)$, also known as the \textit{prior}. As in RL, the agent's world is modelled as a partially observed Markov decision process (POMDP) \citep{Kaelbling1997,Sutton1998,Murphy2000}, where $t$ is the current time, $\tau$ is some timestep in the future and $T$ is the prediction or planning horizon so that $t\leq\tau\leq T$. A policy is a sequence of actions $[a_t, \dots a_\tau, \dots a_T]$ sampled from a Gaussian policy $\boldsymbol{\pi}$ that is conditionally independent across timesteps (i.e., has diagonal covariance matrix). We use the notation $\pi\sim\boldsymbol{\pi}$ for a random policy sample and its corresponding Gaussian policy, respectively. The $\mathrm{FEEF}$ can be separated into an extrinsic (objective-seeking) term and an intrinsic (information-seeking) term when assuming the factorization $\tilde{p}\left(y_\tau, x_\tau\right)\approx q\left(x_\tau\mid y_\tau\right)\tilde{p}\left(y_\tau\right)$:
\begin{equation}
  \label{eq:free_energy_expected_future}
  \begin{aligned}
      \mathrm{FEEF}(\pi)_{\tau} &=\mathbb{E}_{q\left(y_{\tau}, x_{\tau} \mid \pi\right)} \mathrm{D_{KL}}\left[q\left(y_{\tau},x_{\tau}\mid \pi\right)\|\;\tilde{p}\left(y_{\tau},x_{\tau}\right)\right] \\
      &\approx\underbrace{\mathbb{E}_{q\left(x_{\tau} \mid \pi\right)} \mathrm{D_{KL}}\left[q\left(y_{\tau} \mid x_{\tau}\right) \|\; \tilde{p}\left(y_{\tau}\right)\right]}_\text{Extrinsic value }-\underbrace{\mathbb{E}_{q\left(y_{\tau} \mid x_\tau, \pi\right)} \mathrm{D_{KL}}\left[q\left(x_{\tau} \mid y_{\tau}\right) \|\; q\left(x_{\tau} \mid \pi\right)\right]}_\text{Intrinsic value }
  \end{aligned}
\end{equation}
where the difference between the likelihoods $q\left(y_{\tau} \mid x_{\tau}\right)$ and $p\left(y_{\tau} \mid x_{\tau}\right)$ in \autoref{eq:VFE} is simply notational.

Minimizing the extrinsic term biases agents towards policies that yield predictions close to their prior (i.e. their desired future). Maximizing the intrinsic term gives agents a preference for states which will lead to a large information gain -- i.e., the agent tries to visit the states where it will learn the most. The combination of extrinsic and intrinsic value together in a single objective leads to \emph{goal-directed} exploration, where the agent is driven to explore, but only in regions which are fruitful in terms of achieving its goals. There is an additional exploratory factor implicit in the use of a KL-divergence in the extrinsic term, which pushes the agent towards observations where the generative model is not as certain about the likely outcome \citep{Millidge2020c} due to the observation entropy term in the KL-divergence.

The optimal Gaussian policy $\boldsymbol{\pi}^*$ is found through the optimization
\begin{equation}
  \boldsymbol{\pi}^* = \arg\min_{\boldsymbol{\pi}} \sum_{\tau=t}^T\mathrm{FEEF}(\pi)_{\tau}
\end{equation}
by means of a maximum likelihood approach. Although the $\mathrm{FEEF}$ is not a likelihood, \cite{Whittle1991} shows that treating a path integral of a cost as a negative log-likelihood to minimize is formally equivalent to least squares optimization methods, but more direct to compute.

\section{The Active inference capsule}
\begin{figure}[!ht]
  \centering
  \includegraphics[width=0.9\linewidth]{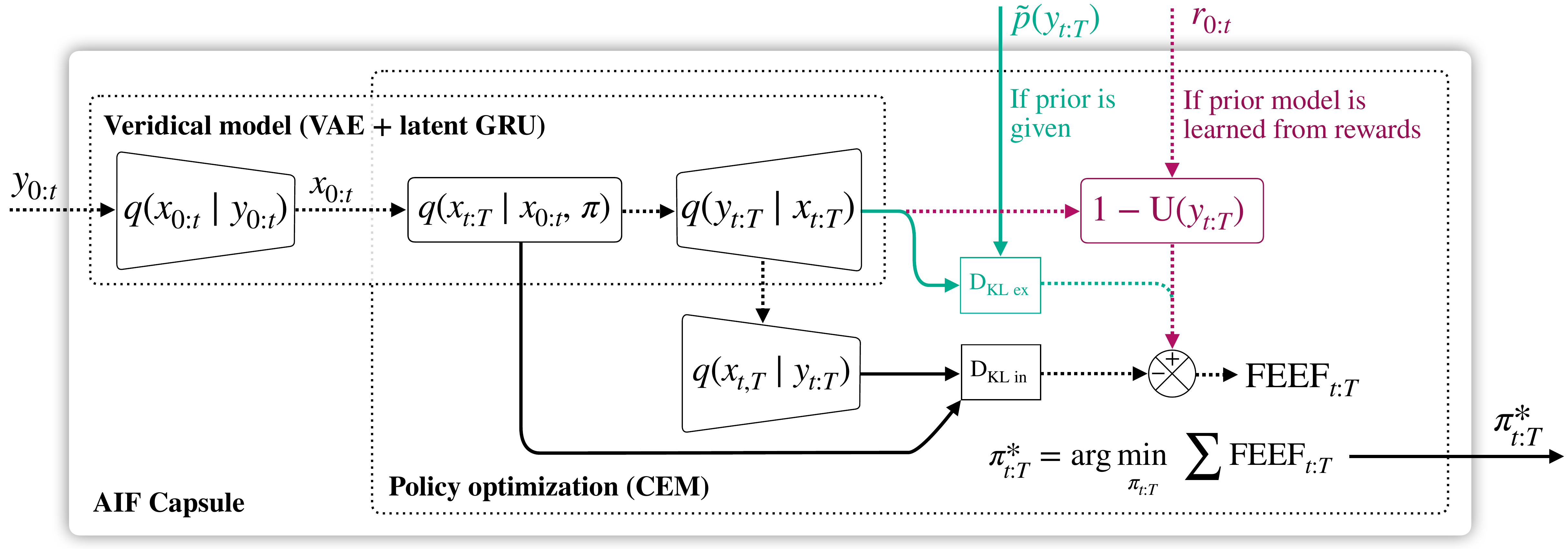}
  \caption{Block-diagram of the active inference capsule. The inputs are time-series of the observations plus the biased priors on future states if given or else the rewards to learn from. The differences between either source of priors are indicated with distinctive colors. The output is a time-series of optimized beliefs over actions (i.e., a Gaussian policy) up to the planning horizon. Continuous lines carry probability distributions whereas discontinuous lines carry real numbers (random samples).}
  \label{fig:AIF_capsule}
\end{figure}

The active inference capsule consists of a variational autoencoder (VAE) which maps the agent's noisy observations to a latent representation, a gated recurrent unit (GRU) \citep{Cho2014} which predicts future latent states from the current latent state, and a policy optimization scheme that minimizes the $\mathrm{FEEF}$ over a trajectory. The VAE and GRU learn an unbiased latent dynamical model in a similar fashion as \textit{world models} by \cite{Ha2018}. Additionally, we propose an extension where the prior model is also learned by the agent from the reward signal. A block-diagram of the capsule is shown in \autoref{fig:AIF_capsule}.

\begin{figure}[!ht]
  \centering
  \includegraphics[width=0.65\linewidth]{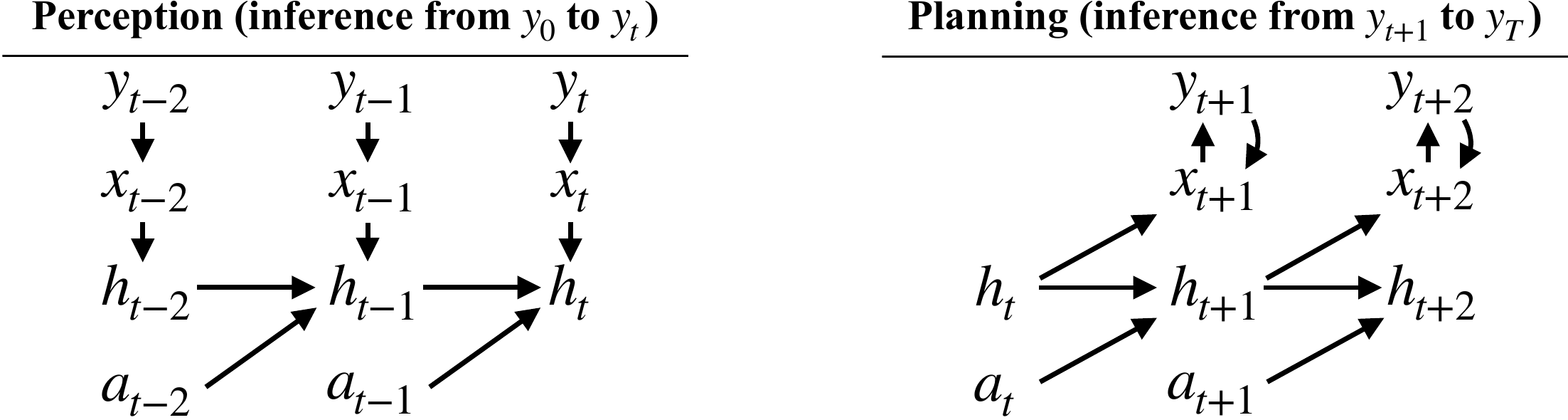}
  \caption{Graphical models of the inference during perception and planning. The future latents are inferred again from the predicted observations, which captures the uncertainty from the variational posterior into the $\mathrm{FEEF}$.}
  \label{fig:graphical_models}
\end{figure}

\begin{figure}[!ht]
  \centering
  \includegraphics[width=0.65\linewidth]{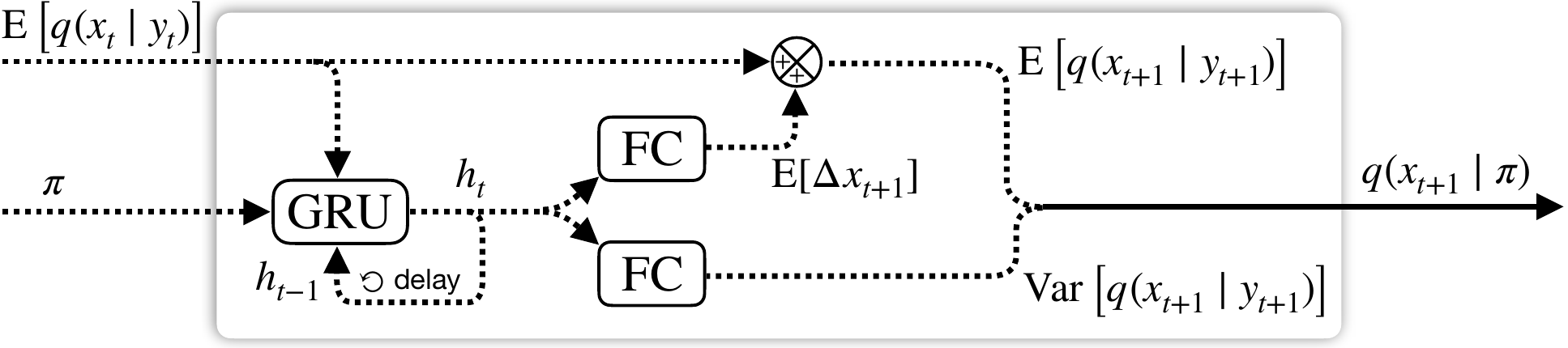}
  \caption{Block-diagram of the transition model.}
  \label{fig:transition_model}
\end{figure}

\paragraph{Perception and planning} Both perception and planning are treated as inference processes (see \autoref{fig:graphical_models}). During perception, the capsule performs inference on the observations $y$ through the unbiased variational posterior and transition models. This updates the belief on the latent states $x$, the recurrent states $h$ of the GRU, which integrate temporal relations between latent states, and the parameters of these models through a learning step. During planning, on the other hand, the current recurrent states are used as initial conditions for the transition model to project future trajectories and evaluate the $\mathrm{FEEF}$ for policy optimization.
\paragraph{Gaussian variational autoencoder} The variational autoencoder (VAE) consists of the variational posterior $q(x_t\mid y_t)$ (encoder) and the likelihood $q(y_t\mid x_t)$ (decoder), both Gaussian and approximated through amortized inference with neural networks \citep{Kingma2014}. As pointed out by \cite{Mattei2018}, the optimization of VAEs for continuous outputs is ill-posed. To circumvent the issue, we fix the variance of the decoder to a default value or to the noise level of the input if given (see the hyperparameters in \autoref{sec:hyperparameters}). The results are only mildly sensitive to this parameter because the minima of the extrinsic term in \autoref{eq:free_energy_expected_future} with respect to the policy only depends on the mode of the likelihood, which is ultimately unaffected by the noise. It does, nonetheless, affect the spread of the likelihood and therefore the gradient of the optimization landscape.
\paragraph{Recurrent transition model} The transition model is factorized into two terms:
\begin{equation}
  q\left(x_{t+1}\mid x_t, \pi\right) \equiv q\left(x_{t+1}\mid h_t\right)q\left(h_t\mid h_{t-1}, x_t, \pi\right)
\end{equation}
The right term is implemented by a GRU which processes the temporal information of the input through a recurrent hidden state $h$. The left term is implemented by a fully-connected (FC) network which predicts both the update on the latent state and the expected variance. Because the recurrent states are deterministic, the variance of the predicted latent distributions does not include prediction errors, which could be an improvement for future work. Our diagram in \autoref{fig:transition_model} differs from \cite{Ha2018} in that it predicts an update on the latent state rather than the latent state itself, as done by \cite{Tschantz2020a}. Learning is achieved via stochastic gradient descent where the loss is the KL-divergence between the predicted latent distribution and the variational posterior after the observation, so that $q\left(x_{t+1}\mid\pi\right)\approx q\left(x_{t+1}\mid y_{t+1}\right)$.
\paragraph{Model free energy}
In contrast to variational autoencoders, which only consider present observations, in an AIF capsule the latent states are also inferred from past sensory data. We propose the following modification of \autoref{eq:VFE} that accounts for the transition model by using it in place of the variational posterior and instead using the variational posterior of the current observation as prior for the predicted latent states:
\begin{equation}
  \label{eq:capsule_free_energy}
    \mathrm{VFE_{capsule}} = \mathbb{E}_{q(x_t\mid x_{t-1}, \pi)}\left[-\ln p\left(y_t\mid x_t\right) \right] + \mathrm{D_{KL}}\left[q(x_t\mid x_{t-1}, \pi)\|q(x_t\mid y_t)\right]
\end{equation}
This expression is used for evaluating the learning progress of the true generative model.

\subsection{Defining the prior}
The AIF capsule supports using both a pre-set prior or else learning one from the reward function. A prior in this context is simply a distribution over goal states. Under active inference, the agent uses its own unbiased world model to generate trajectories that maximize the likelihood of the prior (i.e., reaching the goal). Importantly, when learning the prior the agent can also store information about the optimal solution in a trajectory-independent way by modelling intermediate goal states. Moreover, learning the prior enables the use of active inference agents in situations where manually defining a prior is unfeasible. We propose a novel approach for learning the prior directly based on rewards using Bellman's optimality principle, therefore making a link with model-based reinforcement learning.

\subsubsection{Method for learning the prior model from rewards}
This section presents a novel method for optimal reward shaping \citep{Ng1999} through a continuous potential function that preserves the optimal policy and is compatible with the extrinsic value in \autoref{eq:free_energy_expected_future}. Let $\mathbf{U}(y)$ be a model of the utility of a state $y$ to a trajectory passing through it. The utility is a scaled sum of potential future rewards. We define a reward $r_t$ as a real number in the range $[-1, 1]$ associated with an observation $y_t$, where $-1$ is maximally undesirable and $1$ maximally desirable. We use the same range and definitions for the utility. We suggest a similarity between the utility model and the extrinsic value of the $\textrm{FEEF}$, which instead is a real-valued in the range $[0, \infty)$ where $0$ corresponds to preferences being perfectly fulfilled and $\infty$ to absolutely unpreferred states. Assuming the agent is not too far from its generative model we approximate the relation as
\begin{equation}
  1 - \mathbf{U}(y_t) \underset{\sim}{\propto} \mathbb{E}_{q\left(x_{t} \mid \pi\right)} \mathrm{D_{KL}}\left[q\left(y_{t} \mid x_{t}\right) \|\; \tilde{p}\left(y_{t}\right)\right]
\end{equation}
In other words, both terms have approximately the same landscape. Because the optimal policy does not depend on the absolute values of the $\textrm{FEEF}$ but only on its minima, we can use $1 - \mathbf{U}(y_t)$ during policy optimization as a surrogate for the extrinsic term which implicitly contains the prior.

We model $\mathbf{U}$ as a multi-layered neural network with $\tanh$ activation on the outputs. At every observation, information from the reward signal is infused into the utility model through a stochastic gradient descend (SGD) step. The loss is the mean squared error between the predicted utility and the utility by applying Bellman's equation over a trajectory. In the latter, the discount factor is exponentially decreased for past observations, which pushes the agent to more quickly reach the rewarding states. Moreover, the learning rate in the SGD step is scaled with the absolute value of the reward, which regularizes the magnitude of the model update with the intensity of the stimulus. We also iterate multiple times the process to allow information to propagate back in time more effectively. See \autoref{alg:bellman_learning} for pseudo-code of our dynamic programming approach of learning the utility model.

\begin{algorithm}[!ht]
  \caption{Learning the utility model from rewards}\label{alg:bellman_learning}
  \KwIn{Observations $y_{t_0:t}$ --- rewards $r_{t_0:t}$ --- utility model $\mathbf{U}$ --- discount factor $\beta$ --- learning rate $\alpha$ --- iterations $L$}
  \For{Iteration $i = 1\dots L$}{
    Initialize empty list of expected utilities $\hat{u}$\\
    \For{$\tau=t_0\dots t$ }{
      \eIf{$\tau$ = t}{
        $\hat{u}_t \gets r_\tau$ (in an online setting, future observations are unavailable)\\
      }{
        $\hat{u}_\tau \gets r_\tau + \beta^{t-\tau}\mathbf{U}(y_{\tau+1})$ (Bellman's equation)\\
      }
    }
    $\mathcal{L} \gets \mathrm{MSE}(\mathbf{U}(y_{t_0:t}), \hat{u}_{t_0:t})$ (Compute loss)\\
    $\frac{\partial W_\mathbf{U}}{\partial\mathcal{L}} \gets \mathrm{Backpropagate(\mathbf{U}, \mathcal{L})}$ (Compute weight gradients)\\
    $W_\mathbf{U} \gets W_\mathbf{U} - \alpha\, |r_t|\, \frac{\partial W_\mathbf{U}}{\partial\mathcal{L}}$ (Update weights)
  }
\end{algorithm}

\subsubsection{Policy optimization} Policies are optimized using the cross-entropy method (CEM) \citep{Rubinstein1997}. Since the algorithm is constrained to output Gaussian policies, the exact shape of the $\textrm{FEEF}$ is not captured but the resulting policies do track its minima \citep{Tschantz2020a}. The pseudocode for the optimization is provided in \autoref{alg:cem}.

\begin{algorithm}[!ht]
  \caption{Cross-entropy method for policy optimization}\label{alg:cem}
  \KwIn{Planning horizon $T$ --- Optimization iterations $I$ --- \# policy samples $N$ --- \# candidate policies $K$ --- Transition model $q(x_{t+1}\mid h_t), q(h_t\mid h_{t-1}, x_t, \pi)$ --- encoder $p(x_t\mid y_t)$ --- decoder $p(y_t\mid x_t)$ --- current states $\{x_t, h_{t-1}\}$ --- prior $\tilde{p}(y_t)$}
  
  Initialize a Gaussian policy $\boldsymbol{\pi} \gets \mathcal{N}(\mathbf{0}, \mathbb{I}_{H\times H})$\\
  \For{iteration $i = 1\dots I$}{
    \For{sample policy $j = 1\dots N$}{
      $\pi^{(j)} \sim \boldsymbol{\pi}$\\
      $\textrm{FEEF}^{(j)} = 0$\\
      \For{$\tau = t\dots T-1$}{
        $\mathrlap{h_{\tau}}\hphantom{q(y_{\tau+1} \mid x_{\tau+1})} \gets \mathbb{E}[q(h_\tau\mid h_{\tau-1}, \pi^{(j)}_\tau, x_\tau)]$\\
        $\mathrlap{q(x_{\tau+1}\mid\pi^{(j)})}\hphantom{q(y_{\tau+1} \mid x_{\tau+1})} \gets q(x_{\tau+1}\mid h_\tau)$\\
        $q(y_{\tau+1} \mid x_{\tau+1}) \gets \mathbb{E}_{q(x_{\tau+1} \mid \pi^{(j)})}[q(y_{\tau+1} \mid x_{\tau+1})]$\\
        $q(x_{\tau+1} \mid y_{\tau+1}) \gets \mathbb{E}_{q(y_{\tau+1} \mid \pi^{(j)})}[q(x_{\tau+1} \mid y_{\tau+1})]$\\
        $\mathrlap{\textrm{FEEF}^{(j)}_{\tau+1}}\hphantom{q(y_{\tau+1} \mid x_{\tau+1})} \gets \phantom{-} \mathbb{E}_{q(x_{\tau+1} \mid \pi^{(j)})} \mathrm{D_{KL}}[q(y_{\tau+1} \mid x_{\tau+1}) \|\; \tilde{p}(y_{\tau+1})]$ \\
        $\phantom{q(y_{\tau+1} \mid x_{\tau+1}) \gets}\!-\mathbb{E}_{q(y_{\tau+1} \mid \pi^{(j)})}\mathrm{D_{KL}}[q(x_{\tau+1} \mid y_{\tau+1}) \|\; q(x_{\tau+1} \mid \pi^{(j)})]$\\
        $\mathrlap{\textrm{FEEF}^{(j)}}\hphantom{q(y_{\tau+1} \mid x_{\tau+1})} \gets \textrm{FEEF}^{(j)} + \textrm{FEEF}^{(j)}_{\tau+1}$\\
        $\mathrlap{x_{\tau+1}}\hphantom{q(y_{\tau+1} \mid x_{\tau+1})} \gets \mathbb{E}[q(x_{\tau+1}\mid\pi^{(j)})]$\\
      }
    }
    Select best $K$ policies
    Refit Gaussian policy $\boldsymbol{\pi} \gets \mathrm{refit}(\hat{\pi})$
  }
  \Return{$\boldsymbol{\pi}$}
\end{algorithm}

\section{Experiments on the mountain car problem}\label{sec:experiments}
\makeatletter
\def\blfootnote{\gdef\@thefnmark{}\@footnotetext}
\makeatother
\blfootnote{Code and results at \url{https://github.com/adanielnoel/Active-Inference-Capsule}}
In this section, we study the performance of the active inference capsule using the continuous mountain car problem from the open-source code library OpenAI Gym \citep{Brockman2016}. This is a challenging problem for reinforcement learning algorithms because it requires a substantial amount of exploration to overcome the sparse reward function (negative for every additional action, positive only at the goal). Moreover, the task requires the agent to move away from the goal at first in order to succeed. The objective is to reach the goal in less than 200 simulation steps. In our experiments, the agent time-step size is 6 simulation steps and the planning window $H=T-t$ is defined in the agent's time-scale (see \autoref{sec:implementation} for details).

\paragraph{Online learning} For all tasks, we initialize all the agents with random weights and learn online only. Training an agent for 150 epochs takes about 3 minutes on a single CPU core (Intel I7-4870HQ). In contrast, previous approaches using active inference \citep{Ueltzhoffer2018, Tschantz2019, Tschantz2020a} and policy gradient methods (e.g., \citep{Liu2017}) use (offline) policy replay and typically need hours of GPU-accelerated compute while achieving similar convergence. To our knowledge, this is the first model-based RL method to learn online using neural network representations. This is afforded by the high sample efficiency of the $\textrm{FEEF}$, which directs exploration towards states that are uncertain for both the encoder and transition models.

\begin{figure}[!ht]
  \centering
  \includegraphics[width=0.8\linewidth]{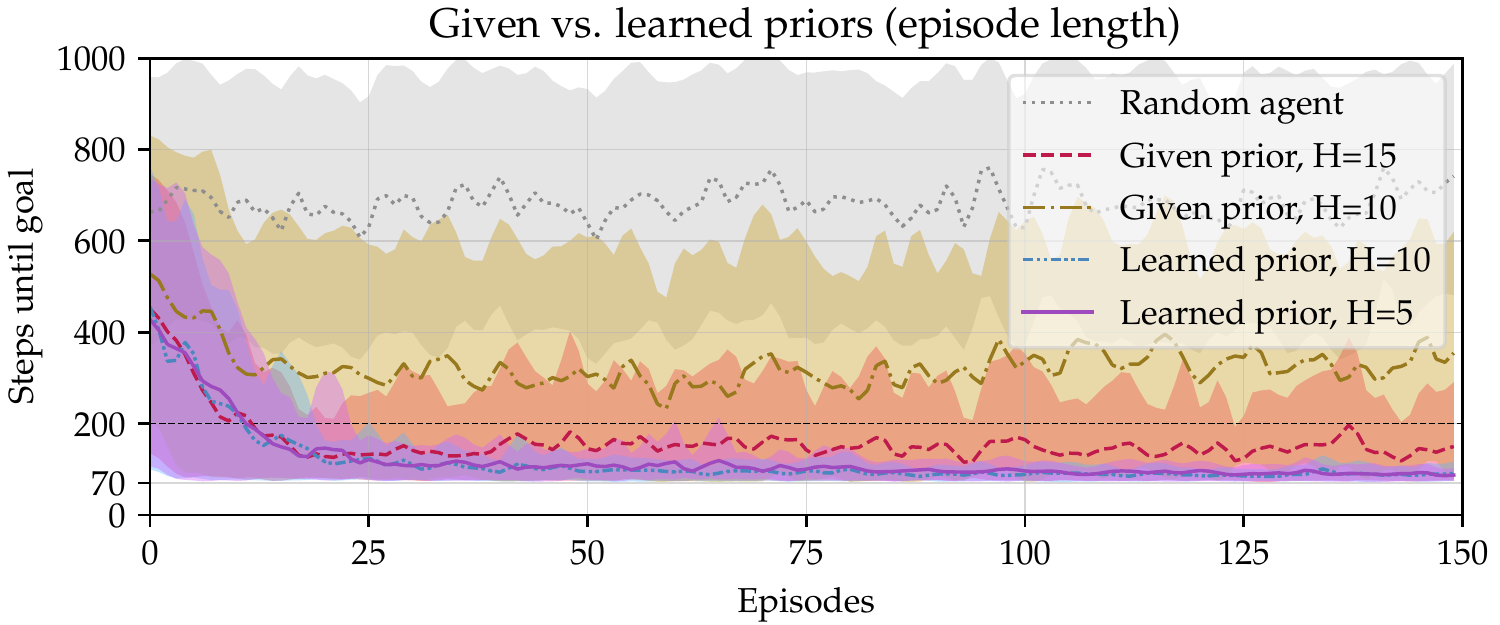}
  \caption{
    Training curves for different types of agents. When given the prior, agents with a planning window of 90 simulation steps (H=15) can reach the goal within the 200-step limit, whereas agents with only a 60-step (H=10) foresight fail. The shortest possible time to the goal is about 70 simulation steps. Agents that learn the prior converge to the optimal solution even if the planning horizon is significantly earlier than 70 steps ahead, showing that the learned prior also captures information about the optimal trajectories and not just the goal. Despite starting from a randomly initialized model, AIF agents can direct exploration already from the first episode, evidenced by the better initial performance compared to agents with purely random actions.}
  \label{fig:goal_vs_rewards_episode_length}
\end{figure}

\begin{figure}[!ht]
  \centering
  \includegraphics[width=1.0\linewidth]{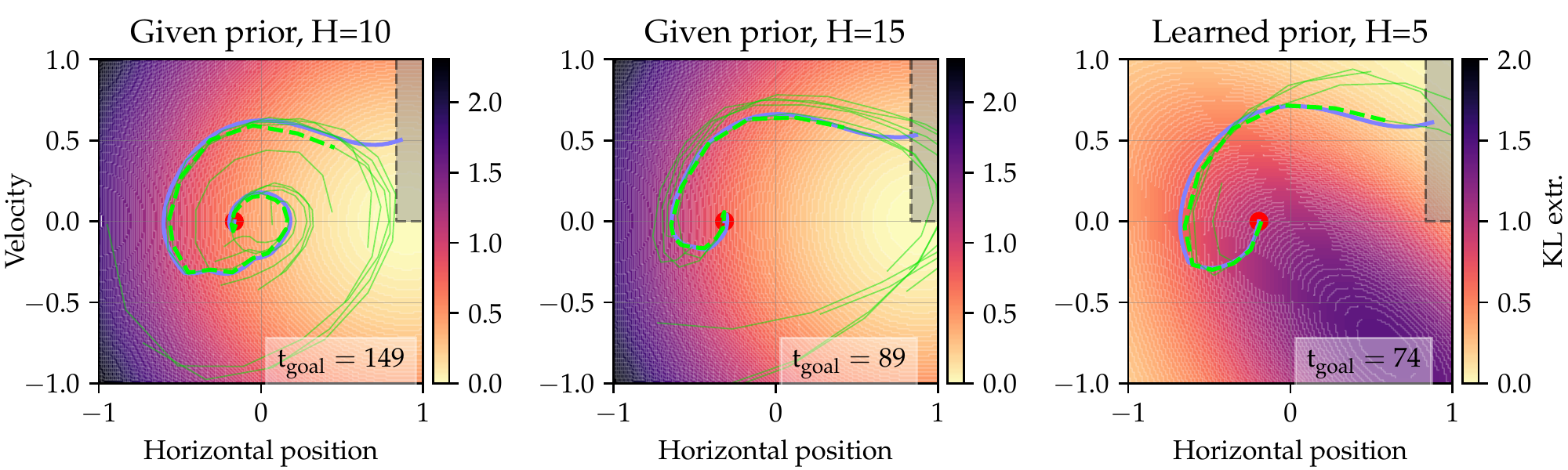}
  \caption{
    Phase portraits of a trained agent of each type. The contour maps are the extrinsic values of the $\textrm{FEEF}$, revealing the Gaussian priors and the learned prior, which also captures information about the optimal trajectory (higher cost in being closer to the goal but having to swing back). The red dots are the initial positions (randomized across trials), the thick continuous lines are the true observations, the thick discontinuous lines are reconstructions by the VAE, and the thin lines are the predictions projected onto the observation space through the decoder model.
  }
  \label{fig:goal_vs_rewards_phase_portraits}
\end{figure}

\paragraph{Given priors versus learned priors} \autoref{fig:goal_vs_rewards_episode_length} shows that agents with a given prior (a Gaussian distribution around the goal state) depend on their planning window to find more optimal policies, whereas agents that learn the prior converge to optimal policies with much shorter planning windows and without such dependency. \autoref{fig:goal_vs_rewards_phase_portraits} shows that the given prior misleads agents with short foresight to swing forward first, whereas the learned prior integrates information about the better strategy and can be followed without a full preview of the trajectory to the goal. These results highlight the importance of the prior for model exploitation. The unbiased predictor is an egocentric model of the world, whereas the prior model is an allocentric representation of the agent's intended behaviour. The active inference capsule effectively combines both during policy optimization, therefore defining a prior based only on the final goal blurs the objective for shorter planning windows. This experiment shows that the reward function is a simple means of indirectly modelling a complex prior.

\paragraph{Exploration properties} \autoref{fig:goal_vs_rewards_episode_length} also reveals that, despite starting without an objective, agents that learn the prior on average find the solution in the first episode faster than agents that take random actions and even agents with a given prior. This is an example of the information-seeking objective of the $\textrm{FEEF}$. It results in a rapid and directed exploration of the state-space, which accelerates the solution to this sparse reward RL problem.

\begin{figure}[!ht]
  \centering
  \includegraphics[width=1.0\linewidth]{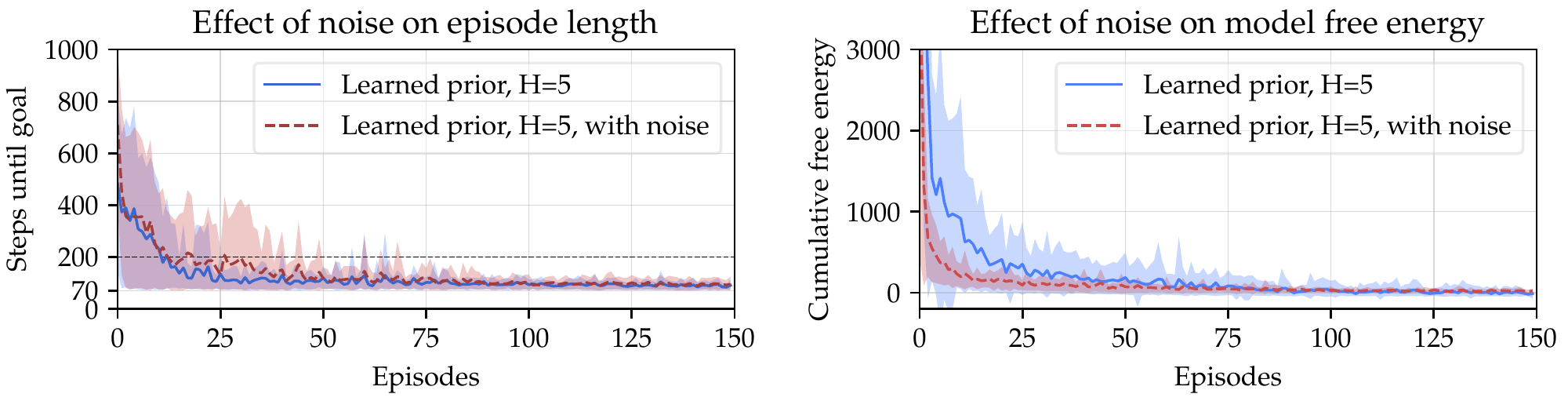}
  \caption{
    \textbf{(Left)} Training curves for agents that learn their own prior, with and without observation noise. Both have very similar convergence, showing that the model is robust to noise. \textbf{(Right)} Cumulative free energy of each episode. The model free energy for agents with observation noise converges much faster, possibly due to the additional regularizing effect against local optima. 
    }
  \label{fig:clean_vs_noise}
\end{figure}

\paragraph{Effect of observation noise} We explore the effect of adding Gaussian noise to the observations. \autoref{fig:clean_vs_noise} shows that, despite a brief initial disadvantage, the agents with noisy observations match the performance of those with clean sensory data and even converge towards the optimal solution. We think that this robustness to observation noise is supported by the KL-divergence in the extrinsic term, as pointed out by \cite{Hafner2020} in the divergence minimization framework. In fact, rather than impairing the capsule, observation noise actually improves learning of the unbiased model, evidenced by the much faster convergence of the model free energy ($\mathrm{VFE_{capsule}}$). \citep{An1996} showed that additional input noise induces a regularizing effect on the backpropagated errors that can improve parameter exploration and prevent overfitting.

\begin{figure}[!ht]
  \centering
  \includegraphics[width=0.75\linewidth]{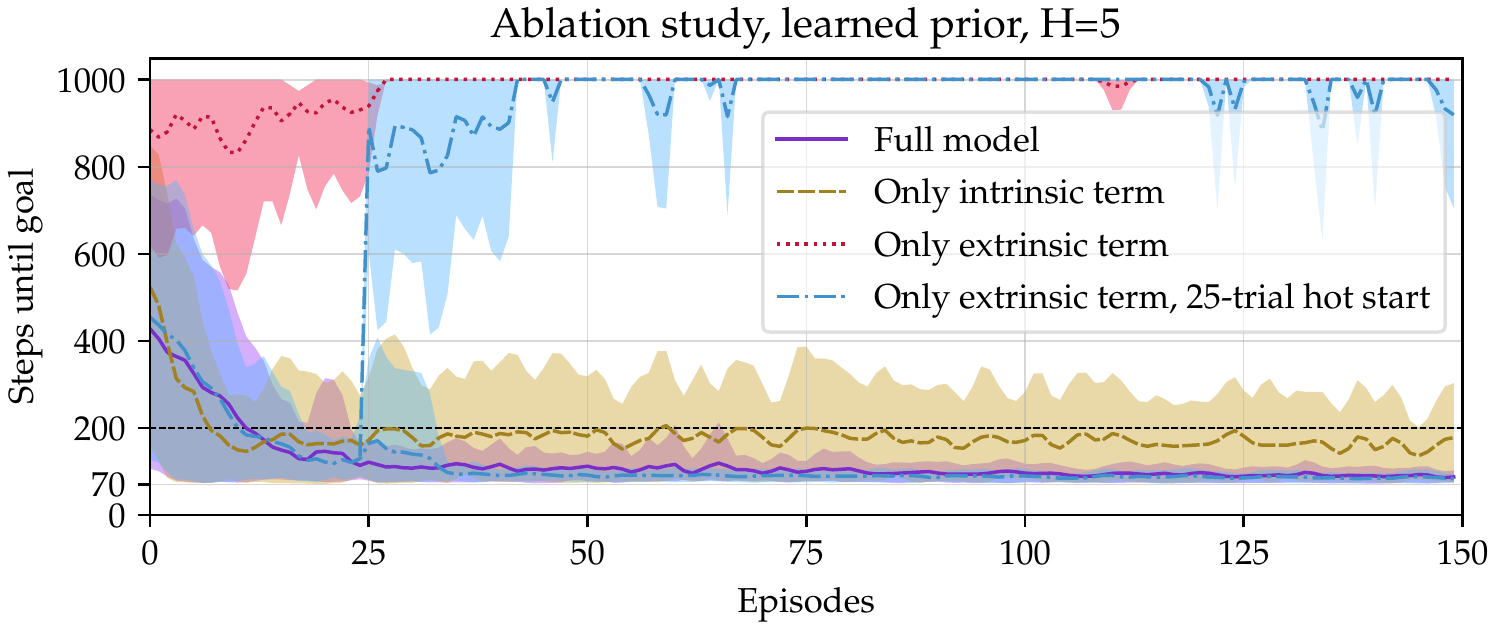}
  \caption{Training curves by selectively disabling the terms of the $\mathrm{FEEF}$. The intrinsic term alone (directed exploration) is enough to solve the mountain car problem. The extrinsic term alone is not enough unless hot-started with some initial episodes with the full model. Then it either overcomes the sparse rewards and converges or it fails to learn a good prior model due to insufficient exploration (the plot is bimodal). The full model combines the convergent but unstable behaviour of the extrinsic term with the robustness furnished by the high sample efficiency of the intrinsic term.}
  \label{fig:ablation_study}
\end{figure}

\paragraph{Ablation study} We explore the contributions of the intrinsic and the extrinsic terms in the behaviour of the agents. \autoref{fig:ablation_study} shows that the intrinsic term alone drives convergent behaviour in the mountain car problem. This is because the goal states are also the rarest (at most once per trial) and therefore directed exploration is both necessary and sufficient to solve the task. Instead, the extrinsic term alone almost never finds the goal state. The extrinsic term promotes exploration of the observation space but not of the latent space (see \autoref{sec:feef}), which results in a lower sample efficiency for model exploration. However, if we hot-start the agent for a few steps before disabling the intrinsic term the behaviour becomes bimodal: the prior model can sometimes gather enough experience for the extrinsic term to maintain a convergent behaviour. These results show that, while the extrinsic term is responsible for the convergence to optimal solutions, the intrinsic term is key for making this behaviour robust because it promotes policies with a high entropy, which prevents convergence to local minima, as well as generates sufficient exploration of the state-space to obtain the sparse reward necessary for learning.

\subsection{Implementation details}\label{sec:implementation}
We update the agent every 6 simulation steps and apply the same action during that period. This simple action-repeat method reduces computation cost, can increase the prediction performance due to higher feature gradients and lower variance of future choices, and is observed in human subjects too \citep{Mnih2015, Cooper-balis2017}. Following the same idea, the agent also commits to executing the first two actions of the chosen policy. Effectively, the agent revises its policy every 12 simulation steps. For each experiment, we train 30 agents of each type and plot their mean and the region of one standard deviation, clipped to the minimum or maximum per episode if exceeded. See \autoref{sec:hyperparameters} for specific details of the neural networks and hyperparameters.

\section{Discussion}\label{sec:discussion}

We introduced an active inference capsule that can solve RL problems with sparse rewards online and with a much smaller computational footprint than other approaches. This is achieved through minimizing the novel \textit{free energy of the expected future} objective with neural networks, which enables a natural exploration-exploitation balance, a very high sample efficiency and robustness to input noise.

Moreover, the capsule can either directly follow a given prior (i.e., the goal states) or learn one from the reward signal using a novel algorithm based on Bellman's equation. We compare both approaches and show that agents with learned priors converge to optimal trajectories with much shorter planning horizons. This is because learned priors approximate a density map of the rewarding trajectories, whereas given priors typically only provide information of the final goals.

Our results show that the $\mathrm{FEEF}$ induces entropy regularization on the policies through uncertainty sampling, which prevents local convergence and accelerates learning. Moreover, our algorithm learns priors that push the agent to achieve its goals as early as possible within the constraints of the problem. The combination of these two characteristics results in a fast and consistent convergence towards the optimal solution in the mountain-car problem, which is a challenge for state of the art RL methods.

Despite the success of the method in the mountain car problem, it remains unclear if the goal-directed exploration properties will scale to high-dimensional inputs or much more complex dynamics. It is also unclear whereas the method we introduced for learning the prior model from the reward signal is generally applicable to any problem. When working with images as inputs, it may be necessary to pre-train the VAE offline on a large dataset and to use a GPU for accelerating computations.

Finally, we believe that the AIF capsule could become a building block for hierarchical RL, where lower layers abstract action and perception into increasingly expressive spatiotemporal commands and higher layers output priors for the lower layers. In this set up, scalability and generality would be achieved by designing wider and deeper networks of AIF capsules, rather than using a large single capsule.

\section*{Broader Impact}\label{sec:broader_impact}
Active inference and the underlying free energy principle describe the self-organising behaviour of biological systems at different spatiotemporal scales, ranging from microscales (e.g., cells), to intermediate scales (e.g., learning processes), to macroscales (e.g., societal organization and the emergence of new species) \citep{Hesp2019}. It is a relatively new science with broad-ranging applications in any technology that has to interact with the real world. But because of its complexity and lack of efficient implementations, active inference has mostly remained an explanatory device with limited applicability outside of the scientific scope. Developments like the active inference capsule presented here may soon unlock the benefits of this new technology for nanobiology, robotics, artificial intelligence, financial technologies, and other high-tech markets.

We acknowledge that the development of this technology may raise safety and ethical concerns in the future, although the scope of the present work is still only methodological. Nonetheless, the model-based nature of active inference renders its decisions partially explainable inasmuch as we understand its priors, which are typically easy to interpret since they are expressed in the observation space (e.g., \autoref{fig:goal_vs_rewards_phase_portraits}). This can be a great advantage over model-free RL methods, which instead are very hard to interpret and to validate for safety-critical applications.

\bibliographystyle{plainnat}
\bibliography{bibliography}

\newpage
\appendix

\section{Implementation details and hyperparameters} \label{sec:hyperparameters}

The variational posterior model has a hidden layer with SiLU activations, which are typically better than ReLU activations in RL settings \citep{Elfwing2018}, and two output layers for mean and standard deviation. The likelihood model has the same structure but outputs a fixed standard deviation (0.05 by default, 0.1 in the case of noisy inputs). The GRU of the transition model has input size $\dim(x)+\dim(a)$ and hidden size $\dim(z)$ parametrized by $2H\cdot\dim(x)$, where $H$ is the planning window in the agent's time-scale. The FC layers map from $\dim(z)$ to $\dim(x)$. The observations are the position and velocity ($\dim(y)=2$) and the actions are the horizontal force ($\dim(a)=1$). The learned prior model consists of a single hidden layer with SiLU activations and Tanh activation on the outputs. The full list of hyperparameters is shown in \autoref{tab:hyperparameters}.

\begin{table}[!ht]
  \centering
  \caption{Agent hyperparameters}
  \label{tab:hyperparameters}
  \begin{tabular}{ll}
    \toprule
    \multicolumn{2}{c}{General hyperparameters} \\
    \midrule
    Latent dimensions $\dim(x)$              & 2                        \\
    VAE hidden layer size                    & 20                       \\
    Observation noise std.                   & 0 or 0.1                 \\
    Time ratio simulation / agent            & 6                        \\
    VAE learning rate (ADAM)                 & 0.001                    \\
    Transition model learning rate (ADAM)    & 0.001                    \\
    \midrule
    \multicolumn{2}{c}{Policy hyperparameters} \\
    \midrule
    Planning window $H$                      & 6, 10 or 15              \\
    Actions before replanning                & 2                        \\
    Policy samples $N$ (CEM)                 & 700 for $H\in\{6, 10\}$  \\
                                             & 1500 for $H=15$          \\
    Candidate policies $K$ (CEM)             & 70                       \\
    Optimization iterations $I$ (CEM)        & 2                        \\
    \midrule
    \multicolumn{2}{c}{Hyperparameters for learned priors} \\
    \midrule
    Hidden layer size (learned priors)       & 40                       \\
    Learning rate (SGD)                      & 0.1                      \\
    SGD steps per reward                     & 15                       \\
    Discount factor $\beta$                  & 0.995                    \\
    \bottomrule                                 
  \end{tabular}
\end{table}

\end{document}